\documentclass{article} 
\PassOptionsToPackage{table,dvipsnames}{xcolor}
\usepackage{iclr2025_conference,times}

\usepackage{amsmath,amsfonts,bm}

\def\eqref#1{equation~\ref{#1}}

\def\1{\bm{1}}

\DeclareMathAlphabet{\mathsfit}{\encodingdefault}{\sfdefault}{m}{sl}
\SetMathAlphabet{\mathsfit}{bold}{\encodingdefault}{\sfdefault}{bx}{n}

\usepackage[utf8]{inputenc} 
\usepackage[T1]{fontenc}    
\usepackage{hyperref}       
\usepackage{url}            
\usepackage{booktabs}       
\usepackage{xcolor}         
\usepackage{amsfonts}       
\usepackage{nicefrac}       
\usepackage{microtype}      
\usepackage{cleveref}

\usepackage{microtype}
\usepackage{graphicx}
\usepackage{makecell}

\usepackage{algorithm}
\usepackage{algpseudocode}
\usepackage{relsize}

\usepackage{amsmath,amssymb,amsfonts,amsthm,mathtools,bm}
\usepackage{enumitem}
\usepackage{multirow}
\usepackage{bm}
\usepackage{array}
\usepackage{color}
\usepackage{wrapfig}
\usepackage{multicol}
\usepackage{diagbox}
\usepackage{pifont}
\usepackage{pdfpages}
\usepackage{kotex}
\usepackage{caption}
\usepackage{subfigure}

\usepackage{algpseudocode}
\usepackage{tcolorbox}

\hypersetup{
  colorlinks,
  linkcolor={red!70!black},
  citecolor={teal},
  urlcolor={blue!70!black}
}

\title{LANTERN++: Enhancing Relaxed Speculative Decoding with Static Tree Drafting for Visual Auto-regressive Models}

\author{
  Sihwan Park\textsuperscript{1$*$} \quad Doohyuk Jang\textsuperscript{1}\thanks{Equal Contribution} \quad
  \bf{Sungyub Kim}\textsuperscript{1} \quad
  \bf{Souvik Kundu}\textsuperscript{2} \quad  
   \bf{Eunho Yang}\textsuperscript{1,3}\thanks{Corresponding Author} \\
\, 
\textsuperscript{1}KAIST  \quad
\textsuperscript{2}Intel Labs  \quad \textsuperscript{3}AITRICS \\
\, 
\texttt{\{sihwan.park, jadohu, eunhoy\}@kaist.ac.kr}\\
\texttt{  sungyub.kim@mli.kaist.ac.kr, souvikk.kundu@intel.com}
}

\iclrfinalcopy 
\begin{document}

\maketitle

\begin{abstract}

Speculative decoding has been widely used to accelerate auto-regressive (AR) text generation. However, its effectiveness for visual AR models remains limited due to token selection ambiguity, where multiple tokens share similarly low probabilities and thus reduce acceptance rates. Recently, relaxed speculative decoding with dynamic tree drafting was proposed to mitigate this ambiguity, demonstrating promising results in accelerating visual AR models. However, we observe that token selection ambiguity still negatively affects dynamic tree drafting, resulting in shallow draft trees and limited acceleration. To overcome this issue, we introduce LANTERN++, a refined framework that integrates static tree drafting with a tailored relaxed acceptance condition, allowing drafts to be selected independently of low-confidence predictions. This enables the acceptance of deeper sequences, improving decoding efficiency while preserving image quality. Extensive experiments on state-of-the-art visual AR models demonstrate that LANTERN++ significantly accelerates inference, achieving up to $\mathbf{\times 2.56}$ speedup over standard AR decoding while maintaining high image quality. The code is publicly available at \url{https://github.com/jadohu/LANTERN}.

\end{abstract}

\section{Introduction}\label{sec:intro}
Recent advances in speculative decoding have significantly accelerated auto-regressive (AR) generation in language models~\citep{leviathan2023fast, medusa, eagle1, eagle2}. However, extending speculative decoding to visual AR models~\citep{llama-gen,chameleon,chern2024anole,liu2024luminamgpt} introduces unique challenges due to the inherent characteristics of image token distributions. In particular, \citet{lantern} identified token selection ambiguity as a major characteristic, where multiple next-tokens share similarly low probabilities. This ambiguity makes it difficult for the drafter model to accurately predict the next token, as even the target model itself lacks confident predictions. As a result, speculative decoding becomes less effective, with drafter predictions often inaccurate and thus frequently rejected by the target model.

Recently, LANTERN~\citep{lantern} was proposed to address this by introducing a relaxed acceptance condition leveraging latent space similarities. Specifically, it allows accepting drafter predictions when they are sufficiently close to the target predictions in latent space, even if they do not perfectly match, motivated by the observation that nearby tokens in latent space yield visually similar outcomes. While LANTERN successfully improved inference speed without compromising image quality, we find that token selection ambiguity still remains problematic within LANTERN's dynamic tree drafting, thereby limiting its full potential. Our analysis reveals two key issues: (1) low drafter confidence results in shallow draft trees, preventing the formation of long accepted sequences, and (2) deterministic selection of draft tokens leads to overly low acceptance probabilities.

To overcome these limitations, we introduce LANTERN++, an enhanced framework that revises LANTERN in two key aspects: (1) the adoption of \textit{static tree drafting} and (2) the integration of a \textit{multiplicative bound} tailored to static tree drafting. By employing static tree drafting, LANTERN++ ensures that deep draft sequences can be generated even under low-confidence predictions. Moreover, static tree drafting allows draft tokens to be selected stochastically rather than deterministically, effectively restoring a normal level of acceptance probability. Furthermore, LANTERN++ refines the relaxation mechanism by replacing the additive distributional distortion bound with a multiplicative bound, which scales acceptance probabilities relative to the original distribution. This adjustment prevents disproportionate favoring of certain tokens, ensuring stable probability scaling and preserving distributional consistency. Extensive experiments demonstrate that LANTERN++ achieves substantial speedups while maintaining high image quality, making it a promising direction for efficient visual AR modeling.

\begin{figure*}
    \vspace{-0.25in}
    \includegraphics[width=1.0\textwidth]{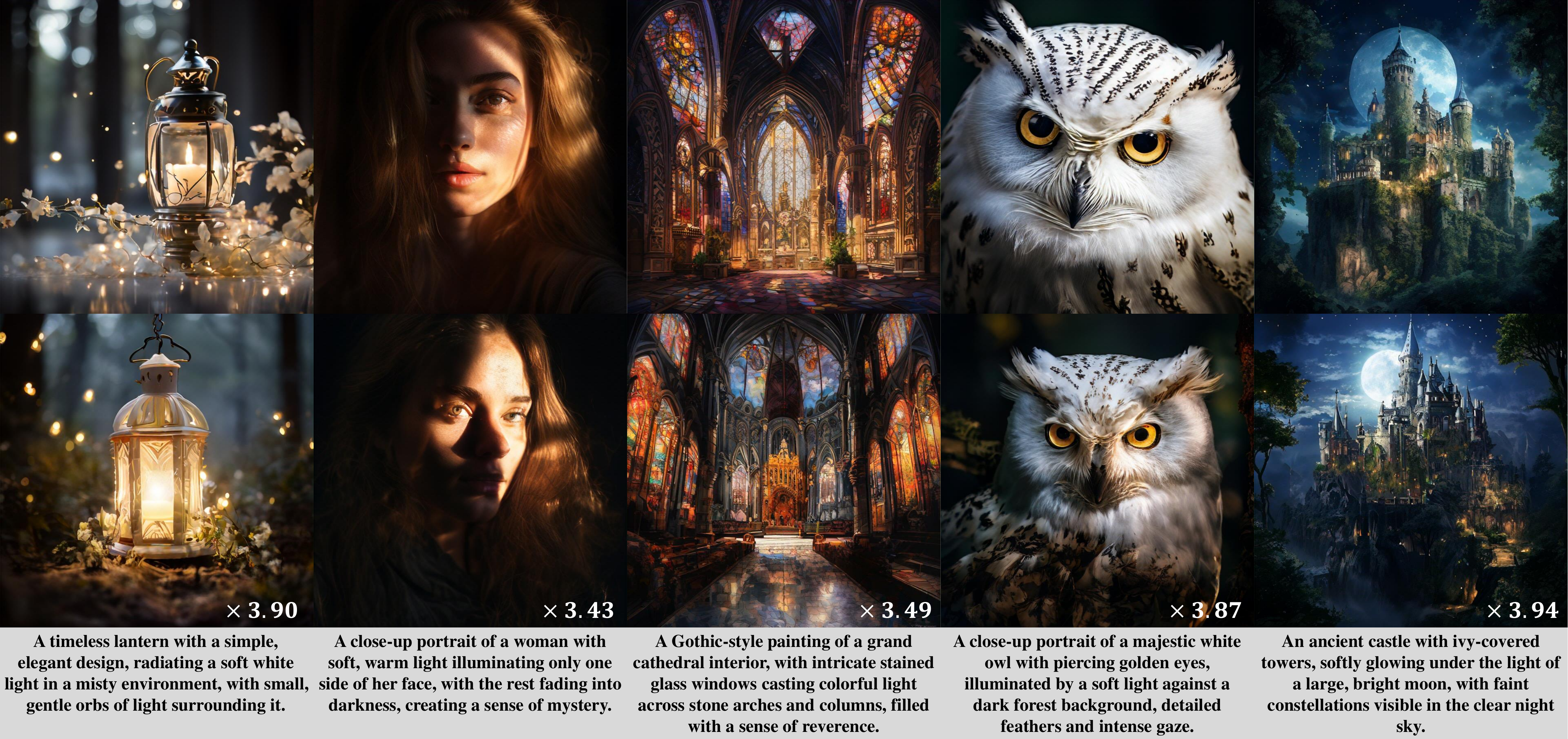}
    \captionof{figure}{Images generated by LANTERN++ on Lumina-mGPT~\citep{liu2024luminamgpt} exhibit remarkable acceleration in decoding steps while maintaining high generation quality. The top row shows images generated by standard auto-regressive decoding, while the bottom row displays images generated with acceleration through LANTERN++. The step compression ratio is presented in white at the bottom right of each image.}
    \label{fig:showcase}
\end{figure*}

Our main contributions are as follows:
\begin{itemize}[leftmargin=1.5em, itemsep=0.5em]
    \item We identify critical limitations of dynamic tree drafting in visual AR models, showing that token selection ambiguity results in shallow draft trees and low acceptance rates, which significantly hamper speculative decoding performance.
    \item We propose LANTERN++, a novel speculative decoding framework that refines LANTERN, which combines static tree drafting with a multiplicative relaxation mechanism, effectively unveiling the full potential of speculative decoding in visual AR models.
    \item We demonstrate through extensive experiments that LANTERN++ achieves up to $\times 2.56$ reduction in latency and $\times 3.63$ step compression across multiple state-of-the-art visual AR models, while maintaining high image quality.
\end{itemize}

\section{Preliminaries}\label{sec:preliminaries}

\paragraph{Notations} We define the \textit{target model} as the visual AR model being accelerated and the \textit{drafter model} as the auxiliary model generating draft tokens. The drafter and target models define probability distributions $p(\cdot|s)$ and $q(\cdot|s)$ where $s$ is the preceding sequence, respectively. Individual tokens are denoted as lowercase $x$ and sequences as uppercase $X$ (e.g., $X_{i:j} = (x_i, \ldots, x_j)$).

\subsection{Tree Drafting and Verification}\label{sec:tree}
Unlike standard chain drafting~\citep{leviathan2023fast}, which generates a single draft token sequence, tree drafting~\citep{medusa,eagle1,eagle2} samples multiple draft tokens per forward pass, forming a tree structure for diverse draft sequences. Given an input \(X_{1:N}\), the drafter first samples \(s_1\) tokens \(\hat{x}_{1,1},\ldots,\hat{x}_{1,s_1}\) from \(p(x|X_{1:N})\), then expands each token with \(s_2\) subsequent drafts in a single forward pass using a tree-aware attention mask. This process continues up to depth \(K\), constructing a draft tree with only \(K\) forward passes. The tree is then flattened and verified by the target model with an appropriate attention mask through a single forward pass.

Tree drafting has two main variants: static tree drafting~\citep{medusa, eagle1} and dynamic tree drafting~\citep{eagle2}. In static tree drafting, the draft tree structure is fixed, where a predefined number of draft tokens is always generated. In contrast, dynamic tree drafting adapts the tree structure based on the drafter model's confidence. Specifically, it calculates a global accept probability $V_i$ for each token $t_i$, defined as: $$V_i = \prod_{t_j \in \text{Path}(\text{root}, t_i)} p_j \approx \prod_{t_j \in \text{Path}(\text{root}, t_i)} c_j,$$ where $p_j$ is the actual accept probability by the target model (unreachable during drafting) and $c_j$ is the drafter's confidence score approximating $p_j$. When expanding the tree, the drafter model generates the top-$k$ candidate tokens for each current leaf node. It then trims the nodes based on their global accept probabilities $V_i$. As a result, dynamic tree drafting produces deeper or shallower draft trees depending on the drafter's confidence at each step.

\subsection{Token Selection Ambiguity and LANTERN}\label{sec:lantern} 
Visual AR models exhibit a phenomenon known as token selection ambiguity~\citep{lantern}. In large language models~\citep{gpt3, touvron2023llamaopenefficientfoundation, jiang2023mistral7b}, the next-token prediction distribution typically forms a distinct, sharp peak, indicating high confidence in a single token choice. In contrast, the next-token distributions in visual AR models are considerably more dispersed. This is primarily due to the continuous, high-dimensional nature of visual tokens (e.g., pixels or image patches), which leads to more uncertain and diffuse next-token prediction distributions. Consequently, it becomes challenging for the drafter model to accurately match the target model's top predictions, substantially limiting the potential of speculative decoding.

To mitigate this ambiguity, LANTERN~\citep{lantern} leverages the observation that tokens located closely in the latent space often represent visually similar and interchangeable semantics. Instead of requiring an exact match with the target model's top prediction, LANTERN relaxes the acceptance condition by aggregating the target probabilities of tokens around the draft token. Intuitively, if a high-probability token predicted by the target model lies near the draft token, aggregating nearby token probabilities can increase the acceptance likelihood of the draft token. However, aggregating probabilities inherently distorts the original target distribution. To avoid excessive distributional shifts, LANTERN further introduces a constraint based on the Total Variation Distance (TVD).

Formally, for a given draft token $\widehat{x}$, LANTERN first identifies its $k$ nearest neighbors in latent space, denoted by $B_k(\widehat{x})$. Then, a TVD bound is applied, yielding a refined neighborhood set $A_{k,\delta}(\widehat{x})\subset B_k(\widehat{x})$ with $\delta>0$ such that the total amount of aggregated probability mass be less than $\delta$. The accept probability for $\widehat{x}$ is then relaxed as $$\text{Original~\citep{leviathan2023fast}: }\min\left(1,\frac{q(\widehat{x}|s)}{p(\widehat{x}|s)}\right)\rightarrow\text{Relaxed: }\min\left(1, \frac{\sum_{x \in A_{k,\delta}(\widehat{x})}q(x|s)}{p(\widehat{x}|s)}\right).$$ If \(\widehat{x}\) is not accepted, resampling is performed based on the adjusted probability distribution. This relaxation increases the effective acceptance rate while preserving visual fidelity, thereby alleviating the issues caused by token selection ambiguity. The more basic preliminaries on visual AR modeling and speculative decoding can be found in the Appendix~\ref{app:prelim}.

\section{Why Dynamic Tree Drafting Fails in Visual AR Models}\label{sec:pitfalls}

Applying dynamic tree drafting to visual AR models, where token selection ambiguity is prevalent, introduces two key issues: (1) draft trees become overly shallow due to low drafter confidence, and (2) the deterministic selection of draft tokens results in consistently low acceptance probabilities. These factors fundamentally limit the effectiveness of speculative decoding by restricting the acceptance of draft sequences, thereby reducing overall acceleration.

\subsection{Shallow Draft Tree}
As described in Section~\ref{sec:tree}, dynamic tree drafting computes a global accept probability for each draft node in the draft tree as the product of the drafter’s confidence scores along the path from the root: $V_i = \prod_{t_j \in \text{Path}(\text{root}, t_i)} c_j$. However, in visual AR models, where drafter confidence scores tend to be low, this product decays exponentially as draft sequence length increases. Consequently, longer draft sequences become significantly less likely to survive, leaving only shorter ones. The result is a draft tree that is wide (many branches) but shallow, preventing the formation of long accepted sequences. Since longer sequences are key to achieving high step compression, this structural limitation restricts the acceleration potential of dynamic tree drafting. An example resulting shallow tree can be found in the Appendix~\ref{app:shallow_example}.

\subsection{Low Accept Probabilities}
In speculative sampling~\citep{leviathan2023fast}, a draft token $\widehat{x}$ is accepted with probability $\min(1, q(\widehat{x}|s) / p(\widehat{x}|s))$, which depends on the alignment between the drafter’s predictions and the target model’s likelihood. However, in dynamic tree drafting, tokens are deterministically selected based on the drafter’s confidence, effectively setting $p(\widehat{x}|s) = 1$. This reduces the acceptance probability to $q(\widehat{x}|s)$. Due to token selection ambiguity, visual AR models produce highly dispersed next-token distributions, with even the most likely token often receiving a probability below 0.2. As a result, even accurate drafter predictions lead to an unnecessarily low acceptance rate.

\begin{table}[t]
    \setlength{\tabcolsep}{10pt}
    \centering
    \renewcommand{\arraystretch}{1.0}
    \caption{Step compression ratios of static (EAGLE-1) and dynamic (EAGLE-2) tree drafting across different modalities. Dynamic tree drafting improves compression in language models (Vicuna-7B~\citep{zheng2023judging}) but performs worse than static tree drafting in visual AR models (LlamaGen-3B~\citep{llama-gen}) due to token selection ambiguity.}
    \label{tab:naive_app}
    \centering
    \resizebox{0.8\linewidth}{!}{
    \begin{tabular}{l|cc}
    \toprule[1pt]
    \multirow{3}{*}{\textbf{Methods}} & \multicolumn{2}{c}{\textbf{Step Compression Ratio}} \\
    \cmidrule{2-3}
    & LlamaGen-3B & Vicuna-7B \\
    \midrule[1pt]
    EAGLE-1 (Static Tree Drafting) & $\times2.49$ & $\times3.94$ \\
    EAGLE-2 (Dynamic Tree Drafting) & $\times2.11$ \textbf{\textcolor{BrickRed}{(-15.3\%)}} & $\times4.98$ \textbf{\textcolor{ForestGreen}{(+26.4\%)}}  \\
    \bottomrule[1pt]
    \end{tabular}
    }
    \vskip -10pt
\end{table}

Table~\ref{tab:naive_app} highlights the contrasting effects of dynamic tree drafting in language models and visual AR models. In language models, dynamic tree drafting improves the step compression ratio from \(\times3.94\) to \(\times4.98\), leveraging concentrated next-token distributions and high-confidence drafter predictions to construct deeper draft trees. In visual AR models, however, the lower drafter confidence produces wide yet shallow draft trees and leads to low acceptance probabilities. Consequently, static tree drafting achieves a higher step compression ratio (\(\times2.49\)) than dynamic tree drafting (\(\times2.11\)), underscoring the latter’s limitations in handling token selection ambiguity.

\section{LANTERN++: Relaxed Acceptance with Static Tree Drafting}
\label{sec:method}

As demonstrated in Section~\ref{sec:pitfalls}, dynamic tree drafting in visual AR models suffers from limited tree depth and low accept probabilities due to token selection ambiguity. Static tree drafting, in contrast, avoids these pitfalls by relying on a fixed tree structure rather than confidence-based adaptive expansion. However, it still struggles with low accept probabilities, as visual AR models inherently assign low probabilities to individual tokens. To address this, we introduce LANTERN++, a refined relaxation of the acceptance condition tailored for static tree drafting.

A key challenge in applying relaxation methods to static tree drafting is the variability of $p(x|s)$. Unlike dynamic tree drafting, where draft tokens are deterministically selected ($p(x|s) = 1$), static tree drafting maintains the true drafter probabilities, which can vary significantly across tokens. This variability makes the original additive relaxation strategy problematic: when $p(x|s)$ is small, an additive increase $\delta$ can lead to disproportionately high accept probabilities, while for larger $p(x|s)$, the boost becomes negligible. This lack of proportionality leads to inconsistent behavior of the relaxation mechanism.

To overcome this issue, we replace the additive relaxation $\delta$ with a \textit{multiplicative distributional distortion bound} $\lambda$, ensuring that accept probability adjustments remain proportional to the target model’s likelihood. Specifically, for each draft token \(\widehat{x}\), its neighborhood \(B_k(\widehat{x})\) is first determined in the same manner as in LANTERN. The selection of refined subset \(A_{k,\lambda}(\widehat{x}) \subseteq B_k(\widehat{x})\) is then changed such that
\[
\text{LANTERN:} \sum_{x\in A_{k,\delta}(\widehat{x})}q(x|s)<\delta\quad \rightarrow 
\quad \text{LANTERN++:} \sum_{x\in A_{k,\lambda}(\widehat{x})}q(x|s) < \lambda\, q(\widehat{x}|s),
\]
where \(\lambda > 1\) is a predefined constant. This ensures that the aggregated target probability does not exceed \(\lambda\) times the original likelihood \(q(\widehat{x}|s)\) (and thus guarantees that the relaxed accept probability does not exceed \(\lambda\) times the original accept probability). The final accept probability is then given by:
\[
\min\left(1, \sum_{x\in A_{k,\lambda}(\widehat{x})}\frac{q(x|s)}{p(\widehat{x}|s)}\right).
\]

This multiplicative relaxation offers two key advantages. First, it maintains a \textit{consistent proportional adjustment} across different values of \(p(\widehat{x}|s)\), ensuring that the acceptance probability is neither excessively inflated nor suppressed. Second, it prevents \textit{instability in cases when \(p(\widehat{x}|s)\) is small}, as a fixed additive boost could lead to over-amplification. By contrast, the multiplicative bound scales naturally with \(q(x|s)\), yielding a more stable and controlled relaxation.

In summary, LANTERN++ improves upon LANTERN through two key components: (1) the use of static tree drafting, and (2) the introduction of a multiplicative relaxation $\lambda$. Static tree drafting enables the robust construction of deep draft trees even under severe token selection ambiguity, and the multiplicative bound further enhances this by providing a stable and consistent relaxation mechanism, complementing the static tree drafting.

\section{Experiments}\label{sec:experiments}

\begin{figure*}
    \includegraphics[width=1.0\textwidth]{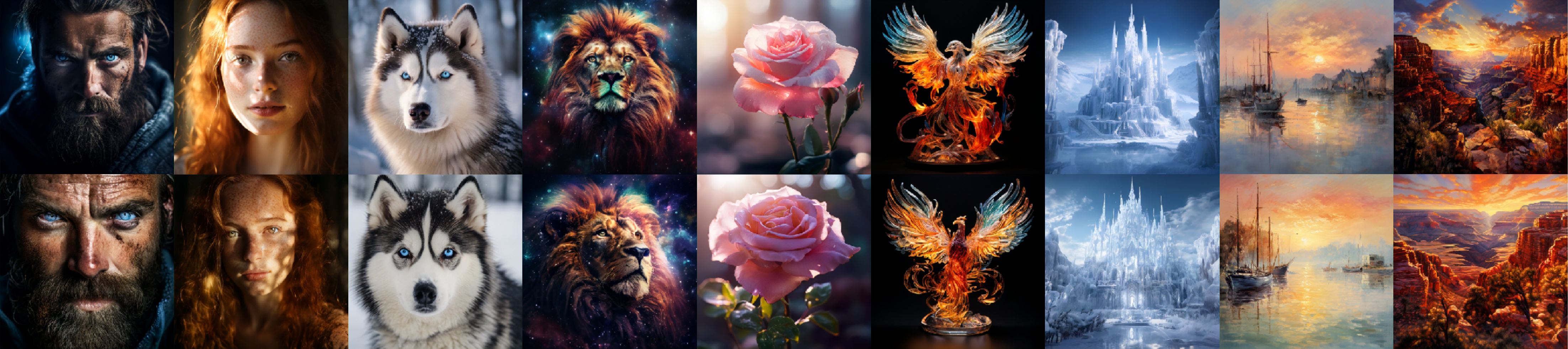}
    \captionof{figure}{Images generated by LANTERN++ on Lumina-mGPT with average step compression ratio $\times 3.63$. The top row shows images generated by standard auto-regressive decoding, and the bottom row displays images generated with acceleration through LANTERN++.}
    \label{fig:qual_samples}
\end{figure*}

We conduct extensive experiments to evaluate the effectiveness of LANTERN++ in accelerating inference for visual AR models while preserving image quality. Our evaluation compares LANTERN++ against standard autoregressive (AR) decoding and EAGLE-1~\citep{eagle1} across multiple state-of-the-art visual AR models, including Lumina-mGPT~\citep{liu2024luminamgpt}, Anole~\citep{chern2024anole}, and LlamaGen-XL~\citep{llama-gen}. All experiments are conducted on the MS-COCO 2017 validation set~\citep{coco}.

For the evaluation of acceleration performance, we measure the step compression ratio $\mathcal{S}=\frac{\text{\# generated tokens}}{\text{\# decoding steps}}$~\citep{fu2024break}, and the actual speed up in terms of latency. Latencies are measured on a highly controlled system to minimize unintended sources of latency. Specifically, we use batch size $1$ on a single A100 PCIe 80GB, with all implementation factors kept constant except for the algorithm. More details on the experimental setting can be found in Appendix~\ref{app:exp_details}.

Table~\ref{tab:main} presents the step compression ratios and inference latency of each method. LANTERN++ consistently outperforms EAGLE-1 across all models, achieving higher step compression and lower latency. Notably, on Lumina-mGPT, LANTERN++ improves the step compression ratio from $\times2.94$ to $\times3.63$, reducing latency from $\times2.10$ to $\times2.56$. Similar trends are observed for Anole ($\times3.41$ vs. $\times2.68$) and LlamaGen-XL Stage 2 ($\times2.86$ vs. $\times2.38$). These results demonstrate that LANTERN++ enables the acceptance of deeper sequences by decoupling token selection from low-confidence predictions, leading to greater acceleration.

In addition to efficiency, we evaluate the impact of LANTERN++ on image quality using the Fréchet Inception Distance (FID)~\citep{fid} and the CLIP score~\citep{clipscore}. While LANTERN++ slightly increases FID due to its relaxed acceptance condition, the overall degradation remains minimal. For example, on Lumina-mGPT, FID increases from $28.93$ (standard AR) to $30.11$ ($\lambda=2$) and $33.91$ ($\lambda=3$), while maintaining a competitive CLIP score. This trade-off highlights the flexibility of LANTERN++ in balancing speed and quality by adjusting the multiplicative bound $\lambda$.

To further assess visual fidelity, Figure~\ref{fig:qual_samples} presents images generated by LANTERN++ vs. standard AR decoding on Lumina-mGPT. Despite the increased acceleration, the visual quality remains \textit{highly consistent}, with no significant loss of detail or artifacts. This qualitative analysis aligns with the quantitative results, confirming that LANTERN++ maintains strong generation quality while substantially reducing inference time. More comprehensive experimental results are provided in Appendix~\ref{app:add_exp}.

\renewcommand{\arraystretch}{1.0}
\begin{table*}[t]
    \centering
    \caption{Acceleration performance and image generation performance of standard decoding, EAGLE-1, and LANTERN++ on various visual AR models using MS-COCO 2017 validation captions.}
    \vskip -2pt
    \resizebox{0.9\textwidth}{!}{
        \begin{tabular}{l|cc|cc}
        \toprule
        \multirow{3}{*}{\textbf{Method}} & \multicolumn{4}{c}{\textbf{MS-COCO 2017 Validation}~\citep{coco}}\\
        \cmidrule{2-5}
        & \multicolumn{2}{c|}{\textbf{Acceleration ($\uparrow$)}} & \multicolumn{2}{c}{\textbf{Image Quality}}\\
        \cmidrule{2-5}
        & Steps & Latency & FID ($\downarrow$) & CLIP Score ($\uparrow$) \\
        \midrule[1pt]
        Lumina-mGPT~\citep{liu2024luminamgpt} & $\times1.00$ & $\times1.00$ & 28.93 & 0.3333 \\
        EAGLE-1~\citep{eagle1} & $\times2.94$ & $\times2.10$ & 29.22 & 0.3325 \\
        LANTERN++ ($k=10, \lambda=2$) & $\times3.19$ & $\times2.28$ & 30.11 & 0.3296 \\
        LANTERN++ ($k=10, \lambda=3$) & $\mathbf{\times3.63}$ & $\mathbf{\times2.56}$ & 33.91 & 0.3267 \\
        \midrule[1pt]
        Anole~\citep{chern2024anole} & $\times1.00$ & $\times1.00$ & 20.28 & 0.3215 \\
        EAGLE-1 & $\times2.68$ & $\times1.75$ & 20.22 & 0.3206 \\
        LANTERN++ ($k=10, \lambda=2$) & $\times2.95$ & $\times1.85$ & 21.10 & 0.3201 \\
        LANTERN++ ($k=10, \lambda=3$) & $\mathbf{\times3.41}$ & $\mathbf{\times2.21}$ & 25.48 & 0.3164 \\
        \midrule[1pt]
        LlamaGen-XL (Stage I)~\citep{llama-gen} & $\times1.00$ & $\times1.00$ & 23.64 & 0.3162 \\
        EAGLE-1 & $\times2.50$ & $\times1.90$ & 23.64 & 0.3162 \\
        LANTERN++ ($k=10, \lambda=3$) & $\mathbf{\times2.98}$ & $\mathbf{\times2.15}$ & 23.89 & 0.3159 \\
        \midrule[1pt]
        LlamaGen-XL (Stage II) & $\times1.00$ & $\times1.00$ & 40.52 & 0.2920 \\
        EAGLE-1 & $\times2.38$ & $\times1.83$ & 40.71 & 0.2925  \\
        LANTERN++ ($k=10, \lambda=3$) & $\mathbf{\times2.86}$ & $\mathbf{\times2.14}$ & 39.80 & 0.2927 \\
        \bottomrule
    \end{tabular}}
    \label{tab:main}
\end{table*}

\section{Conclusion}

We introduced LANTERN++, a speculative decoding framework that significantly improves acceleration in visual AR models by addressing the limitations of dynamic tree drafting in LANTERN caused by token selection ambiguity. Our analysis highlighted the limitations of dynamic tree drafting under token selection ambiguity, where deterministic token selection leads to shallow draft trees and overly conservative acceptance probabilities, reducing acceleration efficiency. By decoupling draft selection from low-confidence predictions and introducing a multiplicative relaxation tailored to static tree drafting, LANTERN++ enables longer accepted sequences and more effective acceleration while maintaining high image quality. However, tuning the multiplicative bound remains crucial for balancing acceleration and image quality. Future work could explore adaptive acceptance mechanisms that dynamically adjust relaxation based on token uncertainty, as well as hybrid drafting strategies that combine the strengths of static and dynamic tree structures to further improve efficiency.

\subsubsection*{Acknowledgments}
This work was partly supported by Institute for Information \&
communications Technology Promotion(IITP) (No.RS-2019-II190075, Artificial Intelligence Graduate School Program(KAIST), No.2022-0-00713, Meta-learning applicable to real-world problems, No.RS-2024-00457882, AI Research Hub Project) and National
Research Foundation of Korea (NRF) (No.RS-2023-
00209060, A Study on Optimization and Network Interpretation Method for Large-Scale Machine Learning) grant funded
by the Korea government (MSIT).

The authors acknowledge the use of ChatGPT (developed by OpenAI) for linguistic assistance, specifically for editing and clarifying the phrasing of sentences and paragraphs. ChatGPT was not involved in any aspect of the research content, including conceptual development, methodological design, or experimental design.

\bibliography{iclr2025_conference}

\begin{thebibliography}{20}
\providecommand{\natexlab}[1]{#1}
\providecommand{\url}[1]{\texttt{#1}}
\expandafter\ifx\csname urlstyle\endcsname\relax
  \providecommand{\doi}[1]{doi: #1}\else
  \providecommand{\doi}{doi: \begingroup \urlstyle{rm}\Url}\fi

\bibitem[Brown et~al.(2020)Brown, Mann, Ryder, Subbiah, Kaplan, Dhariwal, Neelakantan, Shyam, Sastry, Askell, Agarwal, Herbert-Voss, Krueger, Henighan, Child, Ramesh, Ziegler, Wu, Winter, Hesse, Chen, Sigler, Litwin, Gray, Chess, Clark, Berner, McCandlish, Radford, Sutskever, and Amodei]{gpt3}
Tom Brown, Benjamin Mann, Nick Ryder, Melanie Subbiah, Jared~D Kaplan, Prafulla Dhariwal, Arvind Neelakantan, Pranav Shyam, Girish Sastry, Amanda Askell, Sandhini Agarwal, Ariel Herbert-Voss, Gretchen Krueger, Tom Henighan, Rewon Child, Aditya Ramesh, Daniel Ziegler, Jeffrey Wu, Clemens Winter, Chris Hesse, Mark Chen, Eric Sigler, Mateusz Litwin, Scott Gray, Benjamin Chess, Jack Clark, Christopher Berner, Sam McCandlish, Alec Radford, Ilya Sutskever, and Dario Amodei.
\newblock Language models are few-shot learners.
\newblock In H.~Larochelle, M.~Ranzato, R.~Hadsell, M.F. Balcan, and H.~Lin (eds.), \emph{Advances in Neural Information Processing Systems}, volume~33, pp.\  1877--1901. Curran Associates, Inc., 2020.
\newblock URL \url{https://proceedings.neurips.cc/paper_files/paper/2020/file/1457c0d6bfcb4967418bfb8ac142f64a-Paper.pdf}.

\bibitem[Cai et~al.(2024)Cai, Li, Geng, Peng, Lee, Chen, and Dao]{medusa}
Tianle Cai, Yuhong Li, Zhengyang Geng, Hongwu Peng, Jason~D. Lee, Deming Chen, and Tri Dao.
\newblock Medusa: Simple {LLM} inference acceleration framework with multiple decoding heads.
\newblock In \emph{Forty-first International Conference on Machine Learning}, 2024.
\newblock URL \url{https://openreview.net/forum?id=PEpbUobfJv}.

\bibitem[Chern et~al.(2024)Chern, Su, Ma, and Liu]{chern2024anole}
Ethan Chern, Jiadi Su, Yan Ma, and Pengfei Liu.
\newblock Anole: An open, autoregressive, native large multimodal models for interleaved image-text generation, 2024.
\newblock URL \url{https://arxiv.org/abs/2407.06135}.

\bibitem[Chuhmann et~al.(2022)Chuhmann, Köpf, Vencu, Coombes, and Beaumont]{laion-coco}
Christoph Chuhmann, Andreas Köpf, Richard Vencu, Theo Coombes, and Romain Beaumont.
\newblock Laion coco: 600m synthetic captions from laion2b-en, 2022.
\newblock URL \url{https://laion.ai/blog/laion-coco/}.
\newblock September 27th, 2024.

\bibitem[Fu et~al.(2024)Fu, Bailis, Stoica, and Zhang]{fu2024break}
Yichao Fu, Peter Bailis, Ion Stoica, and Hao Zhang.
\newblock Break the sequential dependency of {LLM} inference using lookahead decoding.
\newblock In \emph{Forty-first International Conference on Machine Learning}, 2024.
\newblock URL \url{https://openreview.net/forum?id=eDjvSFOkXw}.

\bibitem[Hessel et~al.(2021)Hessel, Holtzman, Forbes, Le~Bras, and Choi]{clipscore}
Jack Hessel, Ari Holtzman, Maxwell Forbes, Ronan Le~Bras, and Yejin Choi.
\newblock {CLIPS}core: A reference-free evaluation metric for image captioning.
\newblock In Marie-Francine Moens, Xuanjing Huang, Lucia Specia, and Scott Wen-tau Yih (eds.), \emph{Proceedings of the 2021 Conference on Empirical Methods in Natural Language Processing}, pp.\  7514--7528, Online and Punta Cana, Dominican Republic, November 2021. Association for Computational Linguistics.
\newblock \doi{10.18653/v1/2021.emnlp-main.595}.
\newblock URL \url{https://aclanthology.org/2021.emnlp-main.595}.

\bibitem[Heusel et~al.(2017)Heusel, Ramsauer, Unterthiner, Nessler, and Hochreiter]{fid}
Martin Heusel, Hubert Ramsauer, Thomas Unterthiner, Bernhard Nessler, and Sepp Hochreiter.
\newblock Gans trained by a two time-scale update rule converge to a local nash equilibrium.
\newblock In I.~Guyon, U.~Von Luxburg, S.~Bengio, H.~Wallach, R.~Fergus, S.~Vishwanathan, and R.~Garnett (eds.), \emph{Advances in Neural Information Processing Systems}, volume~30. Curran Associates, Inc., 2017.
\newblock URL \url{https://proceedings.neurips.cc/paper_files/paper/2017/file/8a1d694707eb0fefe65871369074926d-Paper.pdf}.

\bibitem[Ho \& Salimans(2021)Ho and Salimans]{cfg}
Jonathan Ho and Tim Salimans.
\newblock Classifier-free diffusion guidance.
\newblock In \emph{NeurIPS 2021 Workshop on Deep Generative Models and Downstream Applications}, 2021.
\newblock URL \url{https://openreview.net/forum?id=qw8AKxfYbI}.

\bibitem[Jang et~al.(2025)Jang, Park, Yang, Jung, Yun, Kundu, Kim, and Yang]{lantern}
Doohyuk Jang, Sihwan Park, June~Yong Yang, Yeonsung Jung, Jihun Yun, Souvik Kundu, Sung-Yub Kim, and Eunho Yang.
\newblock Lantern: Accelerating visual autoregressive models with relaxed speculative decoding, 2025.
\newblock URL \url{https://arxiv.org/abs/2410.03355}.

\bibitem[Jiang et~al.(2023)Jiang, Sablayrolles, Mensch, Bamford, Chaplot, de~las Casas, Bressand, Lengyel, Lample, Saulnier, Lavaud, Lachaux, Stock, Scao, Lavril, Wang, Lacroix, and Sayed]{jiang2023mistral7b}
Albert~Q. Jiang, Alexandre Sablayrolles, Arthur Mensch, Chris Bamford, Devendra~Singh Chaplot, Diego de~las Casas, Florian Bressand, Gianna Lengyel, Guillaume Lample, Lucile Saulnier, Lélio~Renard Lavaud, Marie-Anne Lachaux, Pierre Stock, Teven~Le Scao, Thibaut Lavril, Thomas Wang, Timothée Lacroix, and William~El Sayed.
\newblock Mistral 7b, 2023.
\newblock URL \url{https://arxiv.org/abs/2310.06825}.

\bibitem[Leviathan et~al.(2023)Leviathan, Kalman, and Matias]{leviathan2023fast}
Yaniv Leviathan, Matan Kalman, and Yossi Matias.
\newblock Fast inference from transformers via speculative decoding.
\newblock In \emph{International Conference on Machine Learning}, pp.\  19274--19286. PMLR, 2023.

\bibitem[Li et~al.(2024{\natexlab{a}})Li, Wei, Zhang, and Zhang]{eagle1}
Yuhui Li, Fangyun Wei, Chao Zhang, and Hongyang Zhang.
\newblock Eagle: Speculative sampling requires rethinking feature uncertainty, 2024{\natexlab{a}}.

\bibitem[Li et~al.(2024{\natexlab{b}})Li, Wei, Zhang, and Zhang]{eagle2}
Yuhui Li, Fangyun Wei, Chao Zhang, and Hongyang Zhang.
\newblock Eagle-2: Faster inference of language models with dynamic draft trees.
\newblock \emph{arXiv preprint arXiv:2406.16858}, 2024{\natexlab{b}}.

\bibitem[Lin et~al.(2014)Lin, Maire, Belongie, Hays, Perona, Ramanan, Doll{\'a}r, and Zitnick]{coco}
Tsung-Yi Lin, Michael Maire, Serge Belongie, James Hays, Pietro Perona, Deva Ramanan, Piotr Doll{\'a}r, and C.~Lawrence Zitnick.
\newblock Microsoft coco: Common objects in context.
\newblock In David Fleet, Tomas Pajdla, Bernt Schiele, and Tinne Tuytelaars (eds.), \emph{Computer Vision -- ECCV 2014}, pp.\  740--755, Cham, 2014. Springer International Publishing.
\newblock ISBN 978-3-319-10602-1.

\bibitem[Liu et~al.(2024)Liu, Zhao, Zhuo, Lin, Qiao, Li, and Gao]{liu2024luminamgpt}
Dongyang Liu, Shitian Zhao, Le~Zhuo, Weifeng Lin, Yu~Qiao, Hongsheng Li, and Peng Gao.
\newblock Lumina-mgpt: Illuminate flexible photorealistic text-to-image generation with multimodal generative pretraining, 2024.
\newblock URL \url{https://arxiv.org/abs/2408.02657}.

\bibitem[Loshchilov \& Hutter(2019)Loshchilov and Hutter]{adamw}
Ilya Loshchilov and Frank Hutter.
\newblock Decoupled weight decay regularization.
\newblock In \emph{International Conference on Learning Representations}, 2019.
\newblock URL \url{https://openreview.net/forum?id=Bkg6RiCqY7}.

\bibitem[Sun et~al.(2024)Sun, Jiang, Chen, Zhang, Peng, Luo, and Yuan]{llama-gen}
Peize Sun, Yi~Jiang, Shoufa Chen, Shilong Zhang, Bingyue Peng, Ping Luo, and Zehuan Yuan.
\newblock Autoregressive model beats diffusion: Llama for scalable image generation.
\newblock \emph{arXiv preprint arXiv:2406.06525}, 2024.

\bibitem[Team(2024)]{chameleon}
Chameleon Team.
\newblock Chameleon: Mixed-modal early-fusion foundation models.
\newblock \emph{arXiv preprint arXiv:2405.09818}, 2024.

\bibitem[Touvron et~al.(2023)Touvron, Lavril, Izacard, Martinet, Lachaux, Lacroix, Rozière, Goyal, Hambro, Azhar, Rodriguez, Joulin, Grave, and Lample]{touvron2023llamaopenefficientfoundation}
Hugo Touvron, Thibaut Lavril, Gautier Izacard, Xavier Martinet, Marie-Anne Lachaux, Timothée Lacroix, Baptiste Rozière, Naman Goyal, Eric Hambro, Faisal Azhar, Aurelien Rodriguez, Armand Joulin, Edouard Grave, and Guillaume Lample.
\newblock Llama: Open and efficient foundation language models, 2023.
\newblock URL \url{https://arxiv.org/abs/2302.13971}.

\bibitem[Zheng et~al.(2023)Zheng, Chiang, Sheng, Zhuang, Wu, Zhuang, Lin, Li, Li, Xing, et~al.]{zheng2023judging}
Lianmin Zheng, Wei-Lin Chiang, Ying Sheng, Siyuan Zhuang, Zhanghao Wu, Yonghao Zhuang, Zi~Lin, Zhuohan Li, Dacheng Li, Eric Xing, et~al.
\newblock Judging llm-as-a-judge with mt-bench and chatbot arena.
\newblock \emph{Advances in Neural Information Processing Systems}, 36:\penalty0 46595--46623, 2023.

\end{thebibliography}
\bibliographystyle{iclr2025_conference}

\newpage
\appendix
\section*{Appendix}
\section{Additional Preliminaries}\label{app:prelim}
This section provides additional background necessary for understanding our methods. We adopt notation and exposition styles similar to those presented in \citet{lantern}, with modifications and extensions where appropriate to suit our setting.

\subsection{Visual Auto-Regressive Modeling}

Unlike other image generative models, visual AR models generate images by producing image tokens in an auto-regressive manner. Models used in this study, such as LlamaGen~\citep{llama-gen}, Anole~\citep{chern2024anole}, and Lumina-mGPT~\citep{liu2024luminamgpt}, utilize VQVAE-based discrete tokenizers to decode image tokens into actual images. In the case of text-to-image models, given a tokenized text prompt $T_{1:N}$, image tokens $I_{1:K}$ are generated based on a probability distribution conditioned on the text prompt and the image tokens generated up to the previous step. This can be expressed as:
\begin{align*}
    P(I_{1:K} \mid T_{1:N}) = \prod_{\ell=1}^{K} P(i_{\ell} \mid T_{1:N}, I_{1:\ell-1}),
\end{align*}
where $K$ is determined by the image size, and since each image token is conditioned on the previous image tokens, typical visual AR models require $K$ sequential steps to generate $K$ tokens.

After generating $K$ image tokens, each token is transformed into a visual feature by referencing entries in the codebook $\mathcal{C}$. This codebook, defined as $\mathcal{C} = {c_1, \ldots, c_L}$, contains latent codes $c_i \in \mathbb{R}^d$, where each code serves as a feature vector in the encoder-decoder architecture (e.g., VQVAE or VQGAN), and $d$ represents the feature dimension. Similar to how text tokens correspond to indices, each image token $x_{N+i}$ points to a specific code $c_{x_{N+i}}$ in $\mathcal{C}$. These selected codes are then organized into a tensor of shape $h \times w \times d$ following a raster-scan order (left-to-right, top-to-bottom). Here, $h = H / f$ and $w = W / f$, where $f$ denotes the downsampling factor. The resulting tensor is passed to an image decoder $D: \mathbb{R}^{h \times w \times d} \rightarrow \mathbb{R}^{H \times W \times C}$ to reconstruct the final RGB image.

\subsection{Speculative Decoding}
\paragraph{Draft phase}
Speculative decoding begins with a drafter model that proposes a sequence of $\gamma$ tentative tokens, denoted as $\widehat{X}_{1:\gamma}$, as candidate continuations for an existing input sequence $X_{1:N}=(x_1,\ldots,x_N)$. These tokens are generated autoregressively, where each draft token $\hat{x}_i$ is sampled based on the previously observed tokens and prior drafts according to the distribution $p(\hat{x}_i|(X_{1:N},\widehat{X}_{1:i-1}))$.

\paragraph{Verification phase}
Once the speculative tokens have been produced, the full sequence $(X_{1:N}, \widehat{X}_{1:\gamma})$ is evaluated by a higher-quality target model. Unlike the drafter, the target model processes all draft tokens simultaneously, computing the probabilities $q(\hat{x}_i|(X_{1:N}, \widehat{X}_{1:i-1})$ in a single forward pass. Each token is retained with acceptance probability:
\begin{align}
    \min\left(1, \frac{q(\hat{x}_i|(X_{1:N},\hat{X}_{1:i-1}))}{p(\hat{x}_i|(X_{1:N},\hat{X}_{1:i-1}))}\right).
\end{align}
\label{eq:accept_naive}
Accepted tokens are appended to the output sequence. If a token is rejected, the remaining drafts are discarded, and a replacement for the rejected token is drawn from a normalized distribution based on the positive difference between the target and drafter probabilities:
\begin{align}
    [q(\cdot|(X_{1:N},\hat{X}_{1:i-1}))-p(\cdot|(X_{1:N},\hat{X}_{1:i-1}))]_+
\end{align}
As shown in the analysis by \citet{leviathan2023fast}, this two-stage procedure guarantees that the output sequence remains faithful to the target model’s true distribution.

\paragraph{Tree drafting and decoding} The standard draft and verification phases typically rely on chain drafting (i.e., a single token sequence). However, Medusa~\citep{medusa} and EAGLE~\citep{eagle1, eagle2} introduce a tree structure for draft tokens that significantly improves step compression. Below, we briefly overview tree drafting and its verification.

In tree drafting, each forward pass produces multiple tokens, forming a tree. Given a sequence $X_{1:N}$, the drafter generates $s_1$ tokens $\hat{x}_{1,1},\ldots,\hat{x}_{1,s_1}$ in one pass; for each $\hat{x}_{1,i}$, it then generates another $s_2$ tokens $\hat{x}_{2,i,1},\ldots,\hat{x}_{2,i,s_2}$, and so on, until a limit or maximum depth is reached. A tree-aware attention mask ensures that generating all these tokens still requires only one forward pass per layer of the tree by masking out non-ancestor nodes. Thus, a depth-$K$ tree—containing many more tokens than a chain—can still be built with just $K$ passes.

Tree drafting can be categorized into static tree-based drafting, where the tree structure is predetermined and consistently used, and dynamic tree-based drafting, where the tree structure is dynamically determined based on the drafter's confidence. In LLMs, dynamic tree drafting leverages a well-calibrated drafter to effectively find tree structures that increase the acceptance length, leading to significant speed improvements over static tree-based drafting without requiring additional training for the drafter.

For verification, the draft tree is flattened into a single sequence and fed into the target model with a matching tree-aware mask, allowing verification in a single pass. By widening the range of draft tokens, tree drafting increases the probability of accepting more tokens overall.

\section{Experimental Details}\label{app:exp_details}
\paragraph{Models.}
The evaluation is conducted on three recent visual AR models: LlamaGen~\citep{llama-gen}, Anole~\citep{chern2024anole}, and Lumina-mGPT~\citep{liu2024luminamgpt}. For LlamaGen, we use the text-conditioned LlamaGen-XL (775M) Stage I and II models, which generate images at resolutions of $256\times256$ and $512\times512$, respectively. Anole is a 7B parameter model that generates $512\times512$ images, while Lumina-mGPT is evaluated with the \texttt{Lumina-mGPT-7B-768} variant, which has 7B parameters and generates images at $768\times768$ resolution. In line with EAGLE~\citep{eagle1, eagle2}, each model uses a single decoder layer as the drafter model: a \texttt{LlamaDecoderLayer} for LlamaGen and \texttt{ChameleonDecoderLayer} for both Anole and Lumina-mGPT, with internal dimensionalities matching their respective target models. Note that classifier-free guidance scale is set to be 3.0 and temperature 1.0 and top-$k$ sampling with $k=2000$ is applied for all experiments unless otherwise specified.

\paragraph{Drafter Training.}
For training the drafter model for LlamaGen, we randomly sample 100K images from the LAION-COCO dataset~\citep{laion-coco}. For Anole and Lumina-mGPT, we generate 118K and 30K images using the target model each, based on the captions randomly sampled from the train set of MS-COCO 2017~\citep{coco}. The training procedure is adapted from EAGLE with minor adjustments to accommodate the presence of classifier-free guidance~\citep{cfg}, with a 0.9 probability of selecting text-conditioned samples and a 0.1 probability for null-conditioned samples. AdamW~\citep{adamw} is used as the optimizer with a learning rate $1.0\times10^{-4}$, a batch size of 16, and a total of 20 training epochs.

\begin{figure}
    \centering
    \includegraphics[width=0.9\linewidth]{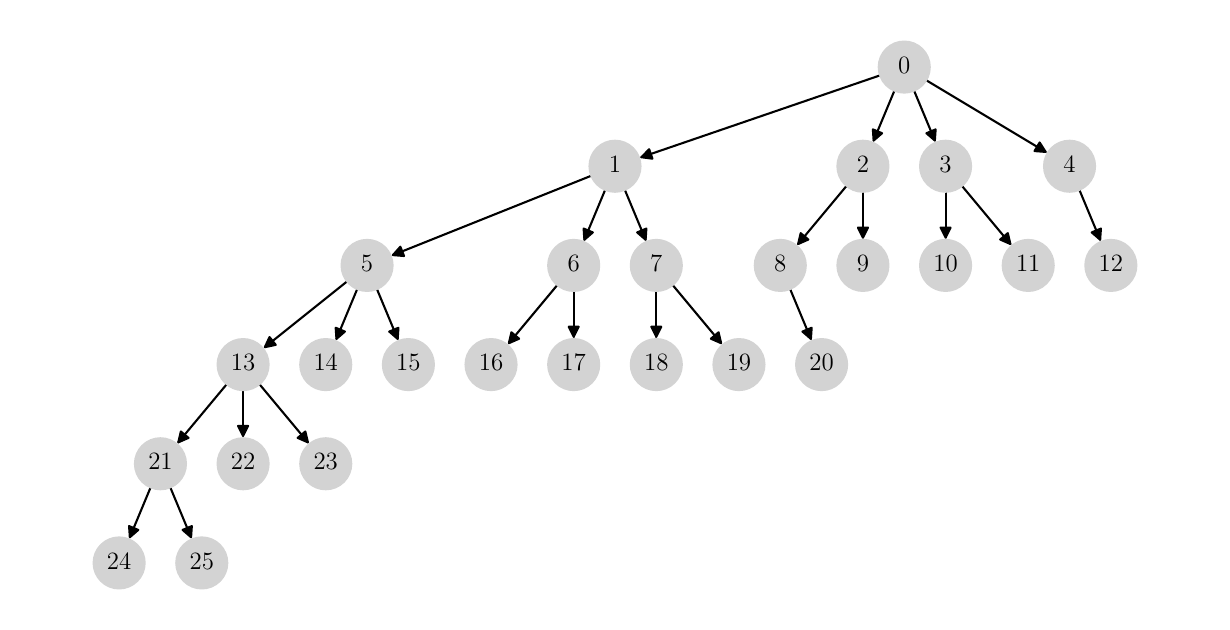}
    \includegraphics[width=0.9\linewidth]{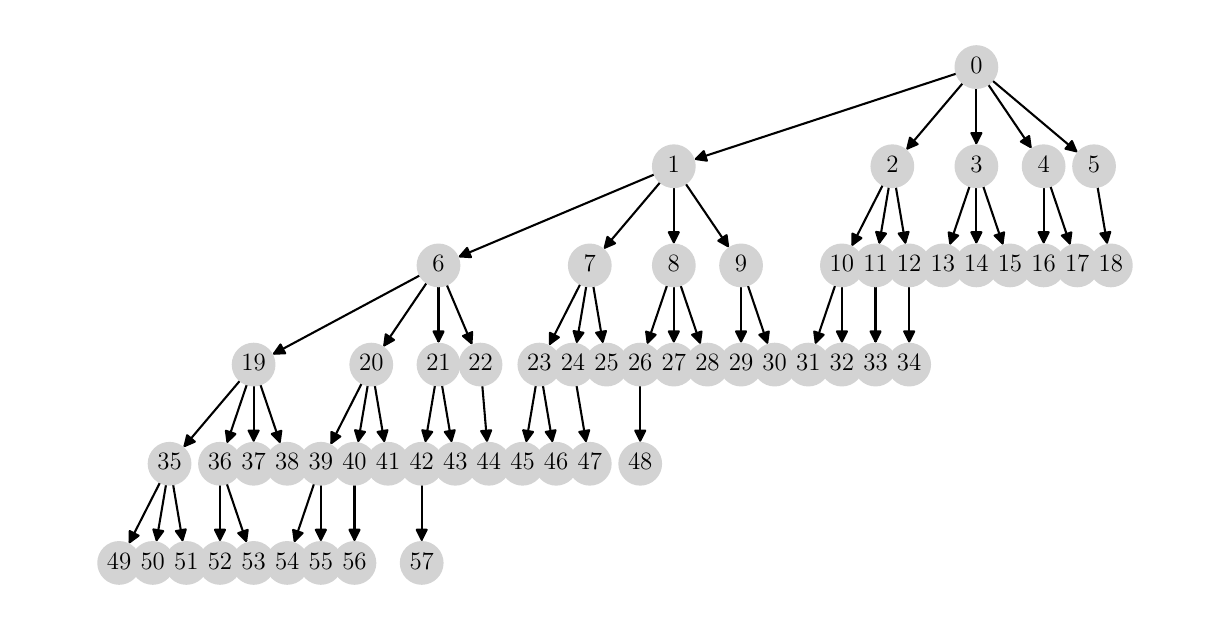}
    \caption{Comparison of static tree structures. \textbf{Top:} The original static tree used in EAGLE-1. \textbf{Bottom:} The extended static tree used for both EAGLE-1 and LANTERN++ in our experiments, designed to match the scale of dynamic tree drafting while maintaining EAGLE-1's structural principles.}
    \label{fig:static_tree}
\end{figure}

\paragraph{Static Tree Structures.}
To ensure a fair comparison with EAGLE-2, which employs dynamic tree drafting with 59 nodes, we extend the static tree structure of EAGLE-1 to include 58 nodes, maintaining a comparable scale. The extension preserves the macroscopic structure of EAGLE-1 while incorporating additional branches to improve depth and coverage. Following EAGLE-1’s design philosophy, we allocate more nodes to the left branches, prioritizing early-stage expansions where acceptance probabilities tend to be higher. As shown in Figure~\ref{fig:static_tree}, this extended static tree structure is used for both EAGLE-1 and LANTERN++ in our experiments to ensure consistency in static tree-based drafting.

\newpage
\section{Additional Experimental Results}\label{app:add_exp}
\subsection{An Example of Shallow Dynamic Tree}\label{app:shallow_example}
\begin{figure}[h]
    \centering
    \includegraphics[width=0.8\linewidth]{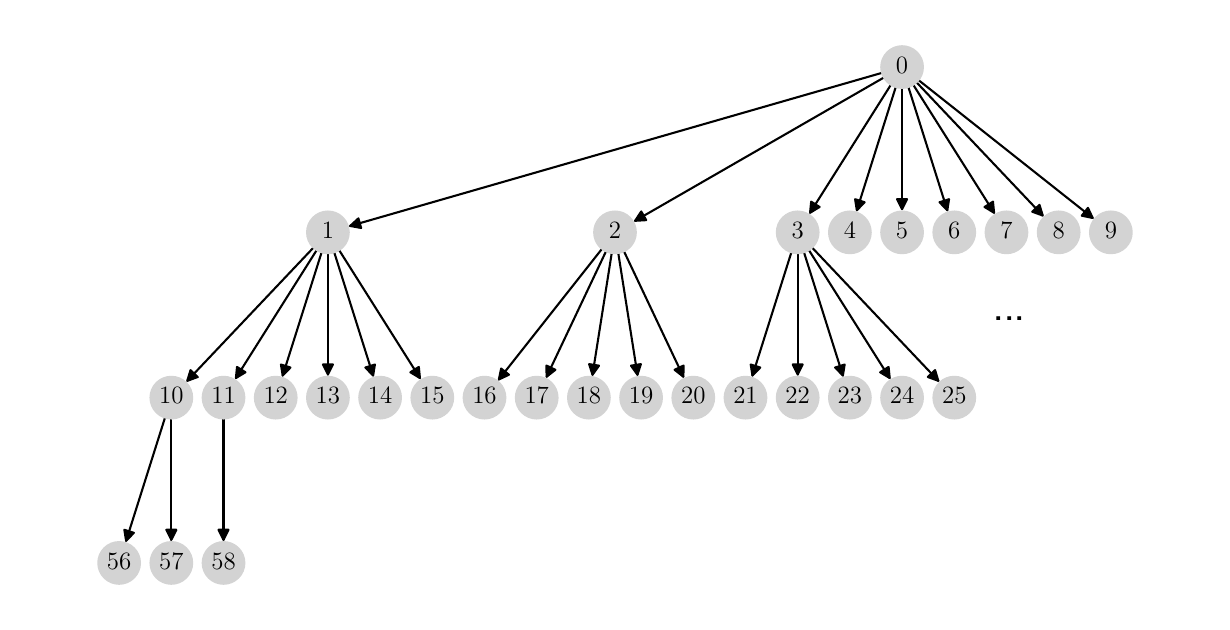}
    \caption{Example of a shallow draft tree produced by dynamic tree drafting. Due to low drafter confidence, deeper expansions are rarely selected, resulting in a tree that is wide but lacks depth, which hinders step compression.}
    \label{fig:shallow_tree}
\end{figure}

Due to token selection ambiguity, dynamic tree drafting often results in a shallow and unbalanced draft tree. As discussed in Section~\ref{sec:pitfalls}, low drafter confidence causes long token sequences to have significantly lower global accept probabilities, leading to a bias toward shorter sequences. Figure~\ref{fig:shallow_tree} illustrates an example tree obtained from dynamic drafting, where most branches remain short, and deeper expansions are rarely selected. This confirms our earlier observation that dynamic tree drafting struggles to construct long accepted sequences, ultimately limiting its acceleration potential.

\subsection{Comparing Additive and Multiplicative Relaxation}

\begin{table}[h]
    \centering
    \caption{Step compression ratios achieved with different values of \(\lambda\) and \(\delta\). Multiplicative relaxation (\(\lambda\)) provides a consistent increase across all settings, whereas additive relaxation (\(\delta\)) exhibits a more gradual improvement due to its uneven effect on tokens with varying probabilities. Results are averaged over 100 randomly sampled captions from the MS-COCO 2017 validation set.}
    \label{tab:relaxation_comparison}
    \resizebox{0.6\linewidth}{!}{
    \begin{tabular}{l|ccc}
    \toprule
    \multicolumn{4}{c}{\textbf{Multiplicative Relaxation (\(\lambda\))}} \\
    \midrule
    \textbf{EAGLE-1: 2.81} & \(\lambda=5\) & \(\lambda=10\) & \(\lambda=20\) \\
    \midrule
    \(k=5\) & 3.02 & 3.17 & 3.30 \\
    \(k=10\) & 3.13 & 3.39 & 3.59 \\
    \(k=20\) & 3.42 & 3.82 & 3.95 \\
    \(k=50\) & 3.68 & 4.13 & 4.46 \\
    \midrule
    \multicolumn{4}{c}{\textbf{Additive Relaxation (\(\delta\))}} \\
    \midrule
    \textbf{EAGLE-1: 2.81} & \(\delta=10^{-4}\) & \(\delta=2\times10^{-4}\) & \(\delta=5\times10^{-4}\) \\
    \midrule
    \(k=5\) & 2.87 & 2.93 & 3.07 \\
    \(k=10\) & 2.83 & 2.96 & 3.35 \\
    \(k=20\) & 2.90 & 3.05 & 3.51 \\
    \(k=50\) & 2.92 & 3.15 & 3.69 \\
    \bottomrule
    \end{tabular}
    }
\end{table}

To evaluate the effectiveness of different relaxation mechanisms in speculative decoding, we compare multiplicative (\(\lambda\)) and additive (\(\delta\)) relaxation in terms of acceleration and image quality. Table~\ref{tab:relaxation_comparison} shows the step compression ratios achieved with varying values of \(k\), \(\lambda\), and \(\delta\). While both methods improve acceleration, a key distinction emerges: \(\lambda\) leads to a more consistent increase across all settings, whereas \(\delta\) exhibits a more gradual improvement with less variation.

This discrepancy arises from how each method interacts with the underlying probability distribution. In multiplicative relaxation, the acceptance probability is scaled in proportion to the original target probability \(q(x|s)\), ensuring that the increase remains relative to the inherent confidence of the target model. This leads to \textit{a stable and uniform acceleration gain}, as every token benefits from a proportional probability boost.

In contrast, additive relaxation applies a fixed probability shift, which interacts differently with tokens of varying probabilities. When a token has a low likelihood, the additive increase has a strong impact, whereas for higher-probability tokens, the effect is relatively weak. Since token probabilities in visual AR models are highly dispersed due to token selection ambiguity, this results in \textit{an uneven relaxation effect across different tokens}. Consequently, \(\delta\) exhibits a more constrained step compression improvement, with some tokens receiving insufficient probability adjustment.

\begin{figure}[h]
    \centering
    \includegraphics[width=0.9\linewidth]{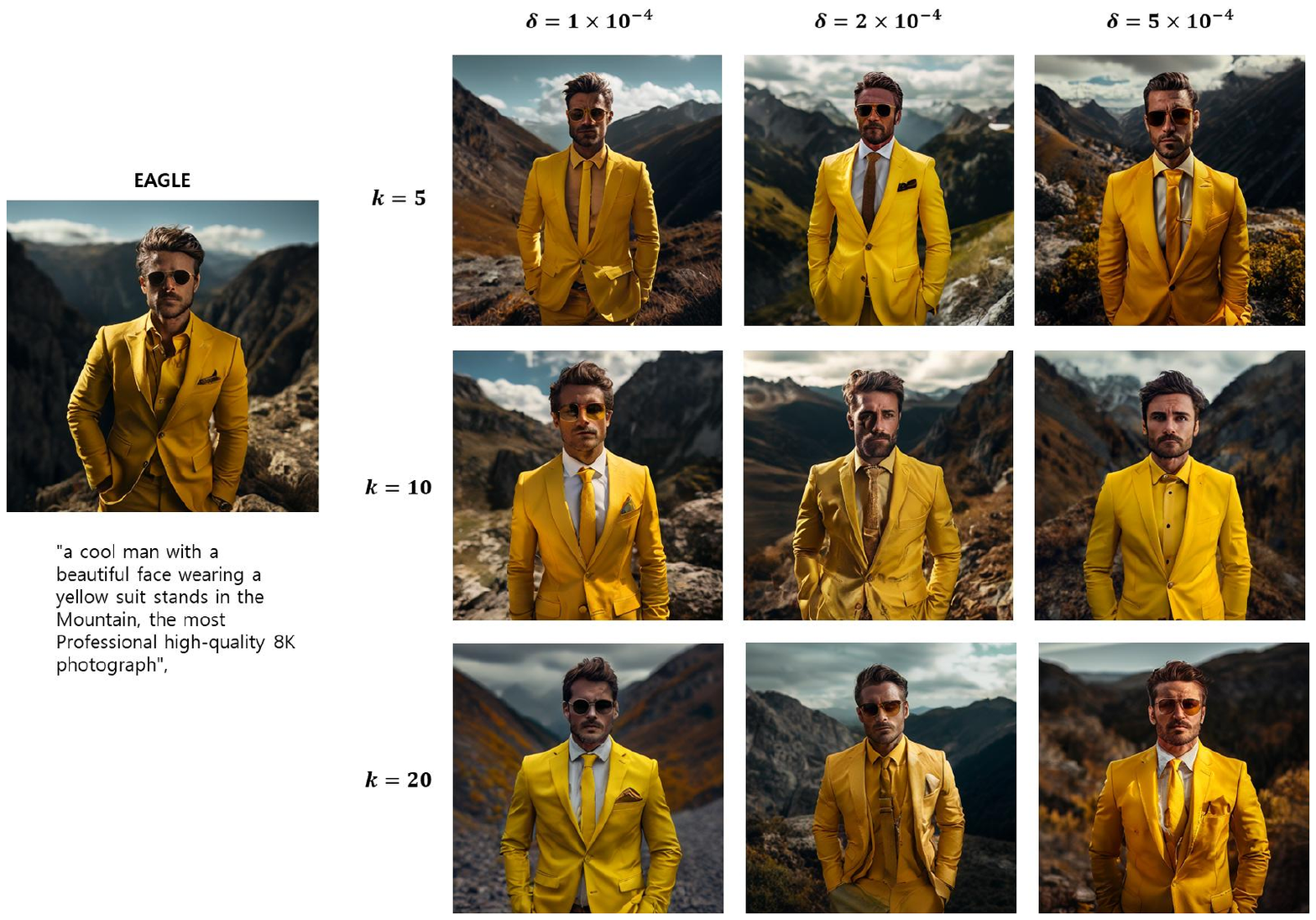}
    \includegraphics[width=0.9\linewidth]{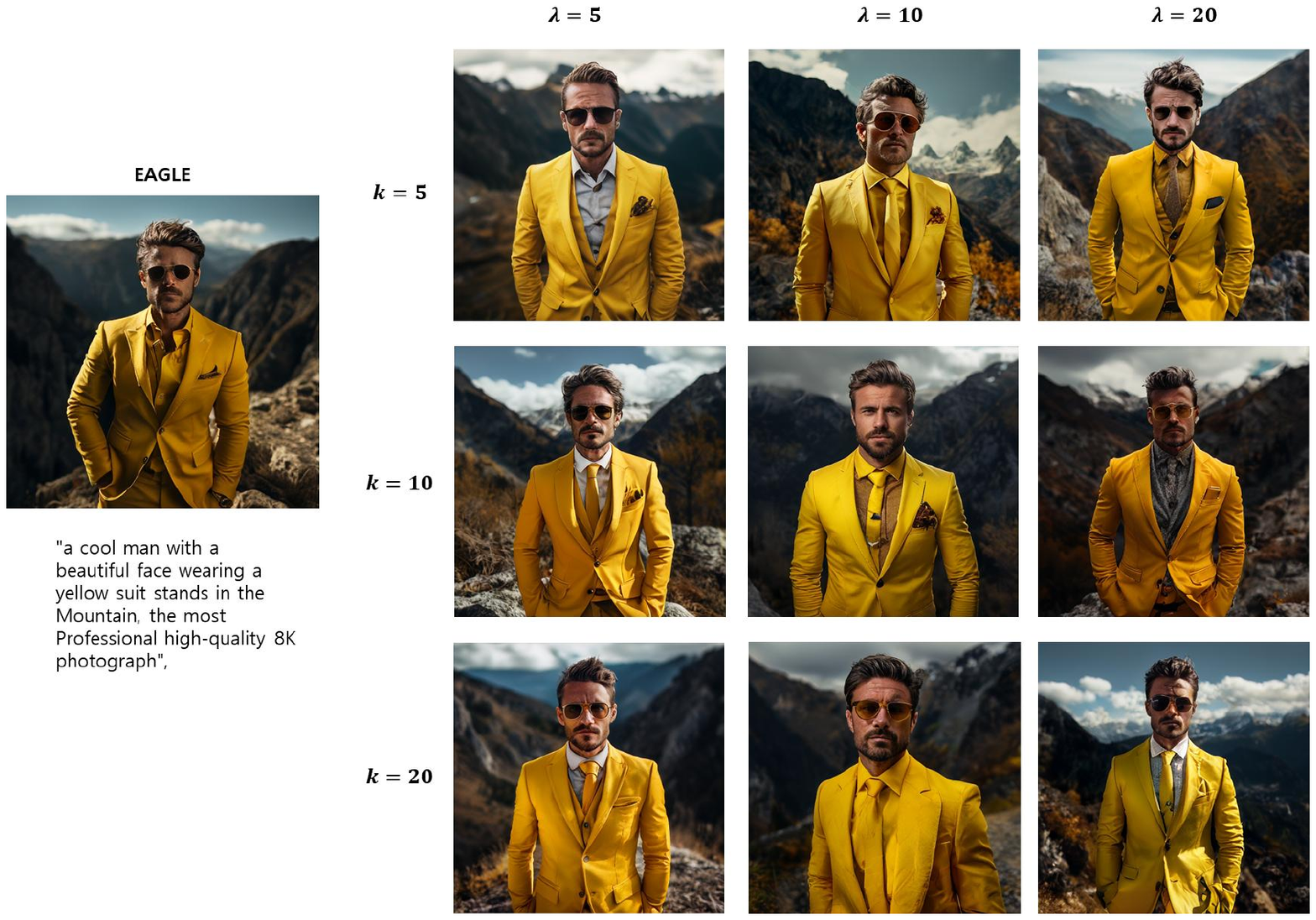}
    \caption{Comparison of images generated using additive relaxation (\(\delta\)) and multiplicative relaxation (\(\lambda\)). The left column shows images generated without relaxation. The middle and right columns show images with increasing levels of \(\delta\) and \(\lambda\), respectively. While both approaches increase acceleration, additive relaxation leads to more noticeable artifacts, whereas multiplicative relaxation maintains stable image quality.}
    \label{fig:relaxation_images}
\end{figure}

The impact of these differences is also visible in image quality, as shown in Figure~\ref{fig:relaxation_images}. While both relaxation strategies maintain reasonable fidelity at lower values, additive relaxation introduces more inconsistencies as \(\delta\) increases. This effect is likely due to uneven probability adjustments leading to local distortions, particularly for tokens that receive disproportionately large or small probability shifts. In contrast, multiplicative relaxation maintains a consistent probability scaling across different token types, resulting in more stable image quality while achieving higher step compression.

Overall, our results suggest that while both approaches can improve speculative decoding efficiency, multiplicative relaxation is better suited for static tree drafting. By ensuring a controlled proportional increase, it provides a predictable balance between acceleration and image preservation, making it a preferable choice for visual AR models.

\newpage
\subsection{Impact of Extended Static Tree on Acceleration and Latency}

\begin{table}[h]
    \centering
    \caption{Step compression ratio comparison between the original and extended static tree. The extended tree consistently improves acceleration across all settings.}
    \label{tab:extended_static_steps}
    \resizebox{0.7\linewidth}{!}{
    \begin{tabular}{l|ccc}
    \toprule
    \textbf{Original (N=26)} & \(\lambda=5\) & \(\lambda=10\) & \(\lambda=20\) \\
    \midrule
    \(k=5\) & 3.02 & 3.17 & 3.30 \\
    \(k=10\) & 3.13 & 3.39 & 3.59 \\
    \(k=20\) & 3.42 & 3.82 & 3.95 \\
    \(k=50\) & 3.68 & 4.13 & 4.46 \\
    \midrule
    \textbf{Extended (N=58)} & \(\lambda=5\) & \(\lambda=10\) & \(\lambda=20\) \\
    \midrule
    \(k=5\) & 3.28 \textcolor{ForestGreen}{(+0.26)} & 3.33 \textcolor{ForestGreen}{(+0.16)} & 3.52 \textcolor{ForestGreen}{(+0.22)} \\
    \(k=10\) & 3.51 \textcolor{ForestGreen}{(+0.39)} & 3.78 \textcolor{ForestGreen}{(+0.39)} & 3.83 \textcolor{ForestGreen}{(+0.24)} \\
    \(k=20\) & 3.63 \textcolor{ForestGreen}{(+0.19)} & 4.03 \textcolor{ForestGreen}{(+0.21)} & 4.24 \textcolor{ForestGreen}{(+0.29)} \\
    \(k=50\) & 3.91 \textcolor{ForestGreen}{(+0.23)} & 4.32 \textcolor{ForestGreen}{(+0.19)} & 4.64 \textcolor{ForestGreen}{(+0.18)} \\
    \bottomrule
    \end{tabular}
    }
\end{table}

\begin{table}[h]
    \centering
    \caption{Latency comparison between the original and extended static tree. The extended tree reduces inference time across most settings despite the increase in context length.}
    \label{tab:extended_static_latency}
    \resizebox{0.8\linewidth}{!}{
    \begin{tabular}{l|ccc}
    \toprule
    \textbf{Original (N=26)} & \(\lambda=5\) & \(\lambda=10\) & \(\lambda=20\) \\
    \midrule
    \(k=5\) & 76.53s & 73.18s & 70.44s \\
    \(k=10\) & 74.26s & 68.69s & 65.05s \\
    \(k=20\) & 68.06s & 61.43s & 59.59s \\
    \(k=50\) & 63.17s & 56.51s & 52.69s \\
    \midrule
    \textbf{Extended (N=58)} & \(\lambda=5\) & \(\lambda=10\) & \(\lambda=20\) \\
    \midrule
    \(k=5\) & 73.50s \textcolor{ForestGreen}{(-3.03s)} & 72.69s \textcolor{ForestGreen}{(-0.51s)} & 68.83s \textcolor{ForestGreen}{(-1.61s)} \\
    \(k=10\) & 68.86s \textcolor{ForestGreen}{(-5.40s)} & 64.67s \textcolor{ForestGreen}{(-3.98s)} & 63.23s \textcolor{ForestGreen}{(-1.82s)} \\
    \(k=20\) & 66.97s \textcolor{ForestGreen}{(-1.09s)} & 60.54s \textcolor{ForestGreen}{(-0.89s)} & 57.24s \textcolor{ForestGreen}{(-2.35s)} \\
    \(k=50\) & 62.36s \textcolor{ForestGreen}{(-0.81s)} & 56.03s \textcolor{ForestGreen}{(-0.48s)} & 52.06s \textcolor{ForestGreen}{(-0.63s)} \\
    \bottomrule
    \end{tabular}
    }
\end{table}

To further improve the acceleration performance of speculative decoding under static tree drafting, we investigate the effect of using an extended static tree with a deeper structure, as illustrated in Figure~\ref{fig:static_tree}. The goal of this extension is to increase the number of draft tokens per step while preserving a balanced trade-off between acceleration and computational efficiency. Tables~\ref{tab:extended_static_steps} and~\ref{tab:extended_static_latency} present the comparison of step compression ratios and latency between the original and extended static tree structures.

The results indicate that increasing the tree depth significantly enhances the step compression ratio. The original static tree (\(N=26\) nodes) achieves a maximum compression of \(\times4.46\) under \(\lambda=20, k=50\), whereas the extended static tree (\(N=58\) nodes) further improves it to \textbf{\(\times4.64\)}, demonstrating a consistent advantage across all \(\lambda\) and \(k\) settings. Notably, the most pronounced improvements occur at moderate \(k\) values (e.g., \(k=10\) and \(k=20\)), suggesting that the extended tree structure provides additional opportunities for token acceptance while maintaining stability in token predictions.

Latency analysis in Table~\ref{tab:extended_static_latency} reveals that despite generating a larger number of draft tokens, the extended static tree \textit{does not introduce significant computational overhead}. In fact, it reduces overall inference latency across most configurations, with improvements reaching up to $5.40$ seconds for \(k=10, \lambda=5\). This suggests that the extended tree structure facilitates speculative decoding by reducing the number of autoregressive decoding steps, thereby lowering the required target model evaluations.

An important consideration is that increasing the number of nodes in draft tree inherently leads to a longer context length during verification, which can introduce computational overhead. However, our results show that despite this potential drawback, the extended static tree consistently achieves faster inference. This improvement arises because the additional draft tokens allow more aggressive speculative decoding, effectively compensating for the increased context length. The reduction in required target model calls outweighs the cost of handling a longer context, leading to an overall speedup.

One interesting observation is that while larger \(\lambda\) values offer higher step compression, their impact on latency reduction diminishes at higher values (e.g., \(k=50, \lambda=20\)). This suggests that while the extended static tree allows for deeper draft sequences, the overall limits of speculative decoding still impose an upper bound on achievable acceleration.

Overall, these results demonstrate that increasing tree depth is an effective strategy for further improving speculative decoding in visual AR models. The extended static tree structure enables higher step compression and lower latency, making it a promising approach for optimizing inference speed in visual AR generation.

\end{document}